\documentclass[9pt, onecolumn]{article}

\usepackage[margin=1in]{geometry}  

\usepackage{cite}
\usepackage{ulem}
\usepackage{xcolor}
\usepackage{acronym}
\usepackage{caption}
\usepackage{comment}
\usepackage{leftidx}
\usepackage{listings}
\usepackage{graphicx}
\usepackage{textcomp}
\usepackage{multirow}
\usepackage{booktabs}
\usepackage{colortbl}
\usepackage{hyperref}
\usepackage{algorithm}
\usepackage{subcaption}
\usepackage{fontawesome5}
\usepackage{algpseudocode}
\usepackage{newtxtext,newtxmath}

\acrodef{TP}{True Positive}
\acrodef{BT}{Behavior Tree}
\acrodef{FP}{False Positive}
\acrodef{FN}{False Negative}
\acrodef{STD}{Standard Deviation}
\acrodef{LLM}{Large Language Model}
\acrodef{FSM}{Finite-State Machine}
\acrodef{ROS}{Robot Operating System}
\acrodef{HRI}{Human-Robot Interaction}
\acrodef{CNN}{Convolutional Neural Network}
\acrodef{SLAM}{Simultaneous Localization and Mapping}

\newcommand{\hl}[1]{\textcolor{blue}{{}{#1}{}}}

\newcommand{\spot}{\raisebox{-0.2em}{\includegraphics[width=1.8em]{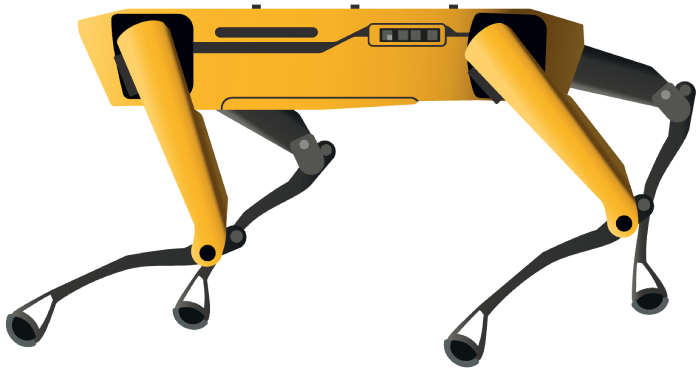}}}
\newcommand{\drone}{\raisebox{-0.2em}{\includegraphics[width=1.8em]{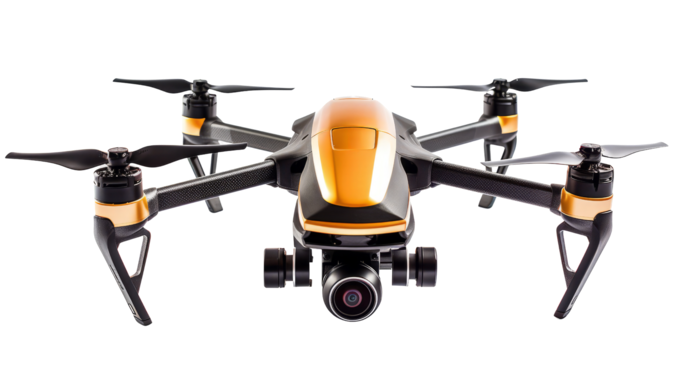}}}

\definecolor{red}{HTML}{fd8f8f}
\definecolor{greend}{HTML}{57e377}
\definecolor{greenl}{HTML}{b8fb8a}
\definecolor{yellow}{HTML}{fefdb4}
\definecolor{orange}{HTML}{ffd5ab}
\definecolor{vividorange}{rgb}{1.0, 0.5, 0.0}

\colorlet{red}{red!50}
\colorlet{yellow}{yellow!50}
\colorlet{greenl}{greenl!50}
\colorlet{greend}{greend!50}
\colorlet{orange}{orange!50}

\begin{document}

\title{Interpretable Robot Control via Structured Behavior Trees and Large Language Models}

\author{
    Ingrid Maéva Chekam$^{1}$, Ines Pastor-Martinez$^{1}$, Ali Tourani$^{1,2}$, Jose Andres Millan-Romera$^{1}$, \\ Laura Ribeiro$^{1}$, Pedro Miguel Bastos Soares$^{1}$, Holger Voos$^{1,3}$, and Jose Luis Sanchez-Lopez$^{1}$
    \thanks{$^{1}$Automation and Robotics Research Group (ARG), Interdisciplinary Centre for Security, Reliability, and Trust (SnT), University of Luxembourg, Luxembourg. (e-mail: \tt{\{ingrid.chekam.001, ines.pastor.001\}@student.uni.lu, pedro.bastos@ext.uni.lu, \{ali.tourani, jose.millan, laura.ribeiro, joseluis.sanchezlopez, holger.voos\}@uni.lu})}
    \thanks{$^{2}$Institute for Advanced Studies (IAS), University of Luxembourg, Luxembourg.}
    \thanks{$^{3}$Faculty of Science, Technology, and Medicine, University of Luxembourg, Luxembourg.}
    \thanks{*This research was funded, in part, by the Luxembourg National Research Fund (FNR), DEUS Project (Ref. C22/IS/17387634/DEUS), RoboSAUR Project (Ref. 17097684/RoboSAUR), and MR-Cobot Project (Ref. 18883697/MR-Cobot). It was also partially funded by the Institute of Advanced Studies (IAS) of the University of Luxembourg through an “Audacity” grant (project TRANSCEND - 2021).}
    \thanks{*For the purpose of open access, and in fulfillment of the obligations arising from the grant agreement, the author has applied a Creative Commons Attribution 4.0 International (CC BY 4.0) license to any  Author Accepted Manuscript version arising from this submission.}
}

\maketitle

\begin{abstract}
As intelligent robots become more integrated into human environments, there is a growing need for intuitive and reliable Human-Robot Interaction (HRI) interfaces that are adaptable and more natural to interact with.
Traditional robot control methods often require users to adapt to interfaces or memorize predefined commands, limiting usability in dynamic, unstructured environments.
This paper presents a novel framework that bridges natural language understanding and robotic execution by combining Large Language Models (LLMs) with Behavior Trees.
This integration enables robots to interpret natural language instructions given by users and translate them into executable actions by activating domain-specific plugins.
The system supports scalable and modular integration, with a primary focus on perception-based functionalities, such as person tracking and hand gesture recognition.
To evaluate the system, a series of real-world experiments was conducted across diverse environments.
Experimental results demonstrate that the proposed approach is practical in real-world scenarios, with an average cognition-to-execution accuracy of approximately $94\%$, making a significant contribution to HRI systems and robots.

The complete source code of the framework is publicly available at \url{https://github.com/snt-arg/robot\_suite}.
\end{abstract}
\section{Introduction}
\label{sec_intro}

With the increasing presence of intelligent robots in everyday life, the demand for reliable and straightforward \ac{HRI} interfaces is rapidly rising.
Traditional robot control paradigms require users to learn particular commands \cite{wang2024hri} or interact with the robots through rigid user interfaces, especially in unstructured environments \cite{fernandez2016natural}.
However, recent works target more flexible and adaptive communication strategies, unlocking the full potential of autonomous agents in human-centered environments.

\begin{figure}[t]
    \centering
    \includegraphics[width=0.9\columnwidth]{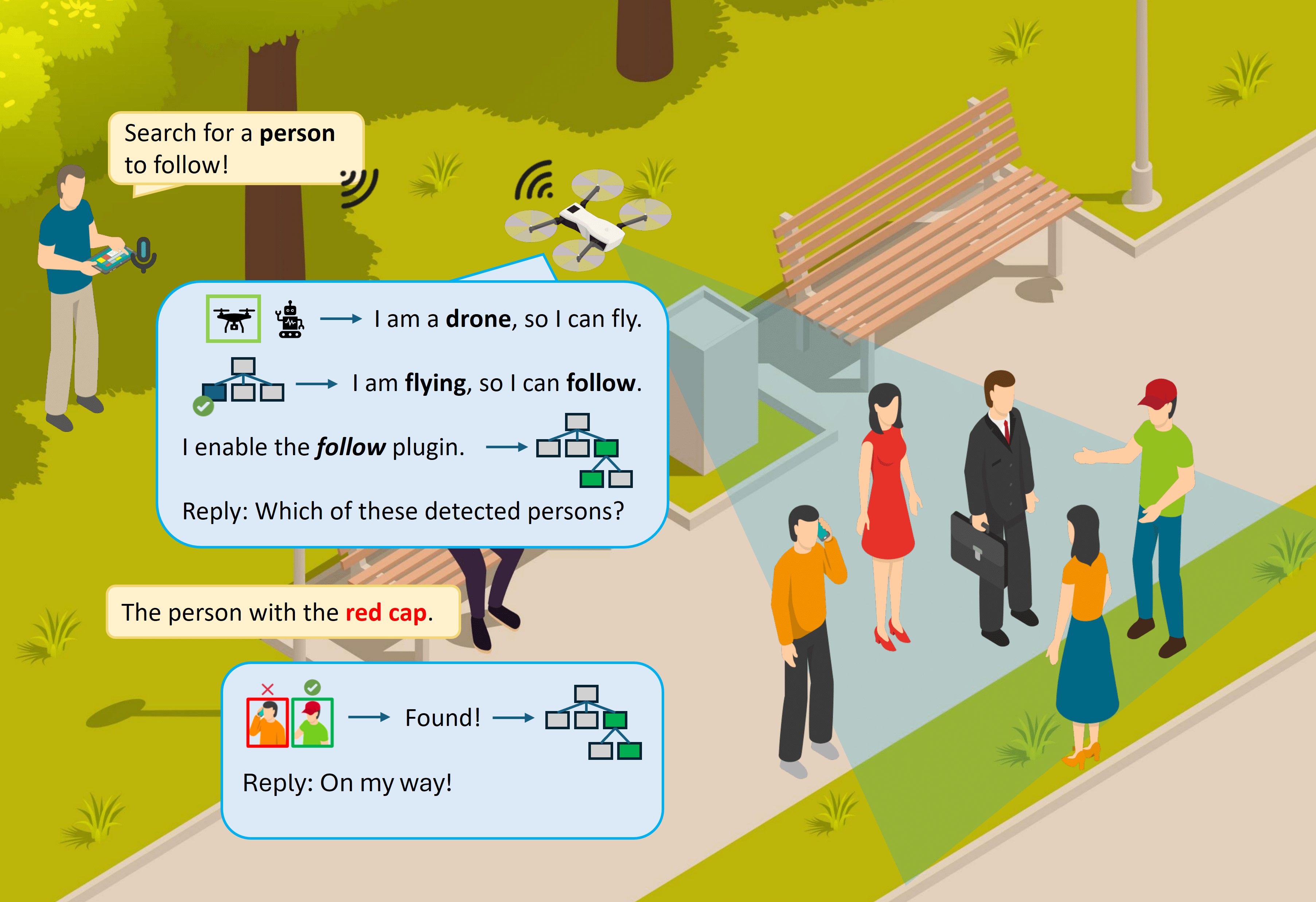} 
    \caption{High-level overview of the proposed LLM-driven robotic control method, where a user interacts with the system through natural language, interpreted by an LLM to guide robot behavior via a modular control structure.}
    \label{fig_overall}
\end{figure}

Accordingly, advances in generative AI and \acp{LLM} reveal new opportunities for enabling seamless communication between humans and robots, where natural language is the primary means of communication \cite{obrenovic2024generative}.
Such models are powerful enough to comprehend given instructions and even ``\textit{reason}'' about the demanded tasks, intentions, and environmental context \cite{kim2024understanding}.
When paired with robotic perception and control systems, \acp{LLM} enable users to intuitively instruct the robot to perform complex tasks such as following multiple objects \cite{hierarchical}, navigating through dynamic scenes \cite{zu2024language}, or interacting with specific items \cite{zhao2024applying}, all using natural dialogue.
Furthermore, integrating multimodal capabilities, including vision and speech, enhances \ac{HRI} by enabling more natural, context-aware communication and improving adaptability across tasks and environments\cite{lami}.

Yet, transforming these high-level instructions into interpretable and reactive robot behaviors requires an underlying control framework.
Various solutions, such as \acp{FSM} \cite{iovino2023programming}, hierarchical planners \cite{cheng2021human}, and learning-based controllers \cite{brunke2022safe}, can aid in structuring robot behavior and decision-making.
However, they often face restrictions in scalability and interpretability when deployed in active, real-world scenarios.
In contrast, \acp{BT} offer a reactive solution for seamlessly incorporating domain-specific modules while preserving a transparent and extensible execution flow \cite{behavtree}.
While \acp{BT} are conceptually similar to \acp{FSM}, they offer superior flexibility, scalability, and clarity, especially for handling complex tasks involving sequencing, fallbacks, or concurrency.
However, their direct manipulation often requires technical expertise, which limits accessibility and hinders intuitive \ac{HRI} for non-expert users.

\begin{figure}[t]
    \centering
    \includegraphics[width=0.6\columnwidth]{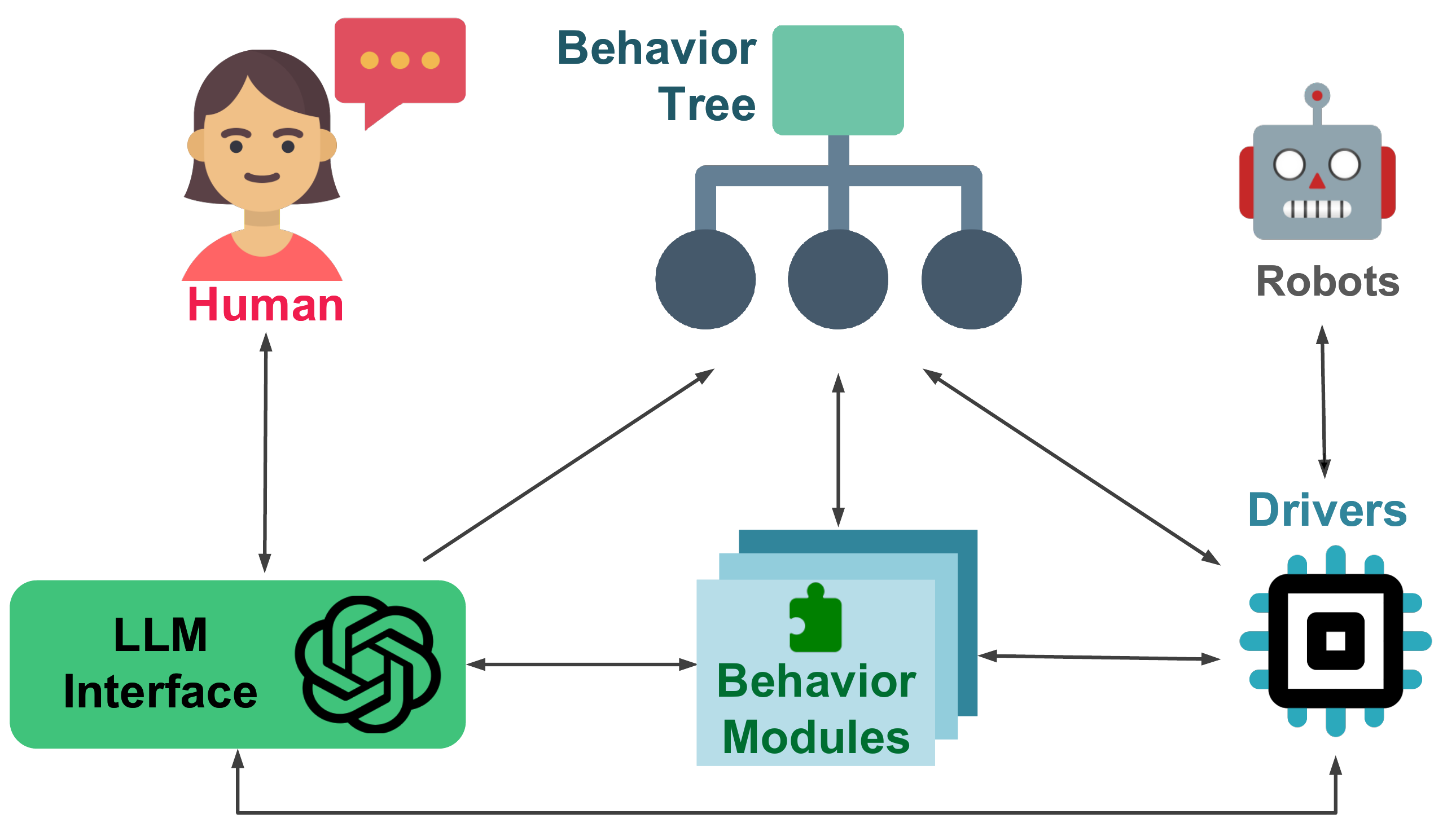}
    \caption{The outline of the proposed system architecture. An LLM interprets natural language instructions from the human, interfacing with a behavior tree to coordinate modular plugins that control the robot’s actions.}
    \label{fig_general_flowchart}
\end{figure}

Bringing these elements together, this paper proposes a novel end-to-end framework that unifies the interpretative power of LLMs alongside the structured execution of BTs to enable robust robot behavior in response to natural language instructions.
Our goal, as illustrated in Fig.~\ref{fig_overall}, is to present an end-to-end framework that supports a dynamic ``command-to-execution'' flow, real-time decision-making, and seamless integration of new behaviors.
In contrast to existing state-of-the-art methods, our approach emphasizes modularity, interpretability, and scalability, by autonomously translating natural language commands into executable actions.
As depicted in Fig.~\ref{fig_general_flowchart}, the \ac{LLM} bridges human objective and actionable robot behaviors by triggering domain-specific behavior modules through the behavior tree.
The system is developed to support multiple behavior modules with effortless integration, primarily within the computer vision domain, including object tracking \cite{kareem2023object} and hand gesture recognition \cite{qi2024computer}.

With this, the paper offers the following contributions:
\begin{itemize}
    \item A modular and model-agnostic robotic framework that integrates LLMs with customized BTs to interpret natural language commands and autonomously execute defined actions,
    \item An open-source implementation of the complete framework, promoting reproducibility and further research; and
    \item Novel systematic evaluation methodologies for measuring success, latency, and robustness of the framework.
\end{itemize}

The remainder of the paper is structured as follows:
Section~\ref{sec_related} reviews similar frameworks that integrate \acp{LLM} and \acp{BT}.
Section~\ref{sec_proposed} details the proposed system and its interior modules.
Experimental results in real-world scenarios are presented in Section~\ref{sec_evaluation}.
Finally, the paper concludes and discusses future directions in Section~\ref{sec_conclusions}.
\section{Related Works}
\label{sec_related}

\subsection{LLM-driven HRI Frameworks}
\label{sec_rw_llm}

Using LLMs for interacting with robots has absorbed significant attention in recent years, enabling more flexible communication between users and robots.
ROSGPT \cite{rosgpt} integrates ChatGPT with robotic systems using prompt engineering and ontologies guidance, though it is limited to single-task execution without autonomous decision-making.
ROS-LLM\footnote{\url{https://github.com/Auromix/ROS-LLM}} \cite{rosllm} generates action plans through ``chain-of-thought'' reasoning and few-shot prompting.
While these plans adopt a tree-like structure for decomposing tasks into sub-actions, they lack core \acl{BT} features, such as hierarchical structuring, fallback mechanisms, and conditional branching.
HELPER \cite{helper} is an embodied agent that uses retrieval-augmented prompting to parse free-form dialogues into action programs, yet remains constrained to simulated settings and abstract tasks despite its memory-based learning.
In contrast, ROSA \cite{rosa2023} offers a scalable solution by integrating a ``reasoning–action–observation'' loop in a flexible agent, where a GPT model autonomously selects tools for both low-level control and high-level reasoning.



Incorporating multimodality, ROSGPT\_Vision\footnote{\url{https://github.com/bilel-bj/ROSGPT_Vision}} \cite{rosgpt_vision} extends ROSGPT by integrating visual language models and perception modules to ground language in a visual context. However, its end-to-end prompt-based design, relying on a single vision module and fixed prompt logic, limits its adaptability to dynamic environments and tasks that require context-sensitive perception.
LaMI \cite{lami} enhances multimodal understanding by integrating speech (audio), vision, text, and gestures to infer high-level human intent.
However, its main limitation lies in its substantial computational overhead, stemming from its intricate multimodal fusion mechanisms.


\subsection{LLMs for Behavior Tree Control}
\label{sec_rw_bt}

Other works integrate \acp{BT} with LLMs to facilitate its use, enabling intuitive behavior control via natural language \cite{tagliamonte2024generalizable}.
Specifically, LLM-BRAIn \cite{llmbrain} and BTGenBot \cite{btgenbot} utilized LLMs to ``generate'' Behavior Trees from natural language instructions.
LLM-BRAIn primarily focuses on model optimization for deployment on lightweight mobile robotic platforms, aiming to minimize computational overhead.
On the other hand, BTGenBot focuses on fine-tuning language models to generate behavior trees with enhanced structural quality.
While both approaches contribute to the automated generation of \acp{BT}, they lack support for interactive modification or real-time interaction during execution, which is critical for adaptive \ac{HRI}.

Apart from generating \acp{BT} using LLMs, some approaches focus on modifying predefined \ac{BT} structures that contain robot behaviors.
In \cite{10409012}, GPT language models are employed to indirectly manipulate XML-driven \acp{BT}, including querying, generating, and modifying its functions.
The primary weakness of this approach is the introduction of an extra layer of complexity and potential integration latency.
BETR-XP-LLM \cite{styrud2024automatic} and LLM-BT \cite{llmbt} enable real-time BT adaptation and modification, minimizing human intervention.
In this regard, BETR-XP-LLM combines a long-horizon planner with an LLM to handle unexpected failures, providing reasoning feedback to operators.
LLM-BT utilizes ChatGPT for reasoning and a BERT-based model to extract keywords; however, its keyword parsing relies on rigid \texttt{if/else} rules, which limit task generality.
Similarly, DiaGBT \cite{diagbt} uses an LLM to define mapping rules on sub-BTs for constructing complex BTs and incorporates \textit{memory} to learn reusable skills.
Thought it requires imperative commands from the user and relies on Q\&As templates for disambiguation, which can hinder the HRI.

It can be seen that the existing works lack a direct, dynamic command-to-execution flow that enables real-time modification of Behavior Trees using an LLM.
The proposed framework aims to address these gaps by introducing a generalizable architecture for seamless integration of LLM-driven interpretation with dynamic behavior tree execution.

\begin{figure*}[t]
    \centering
    \includegraphics[width=\textwidth]{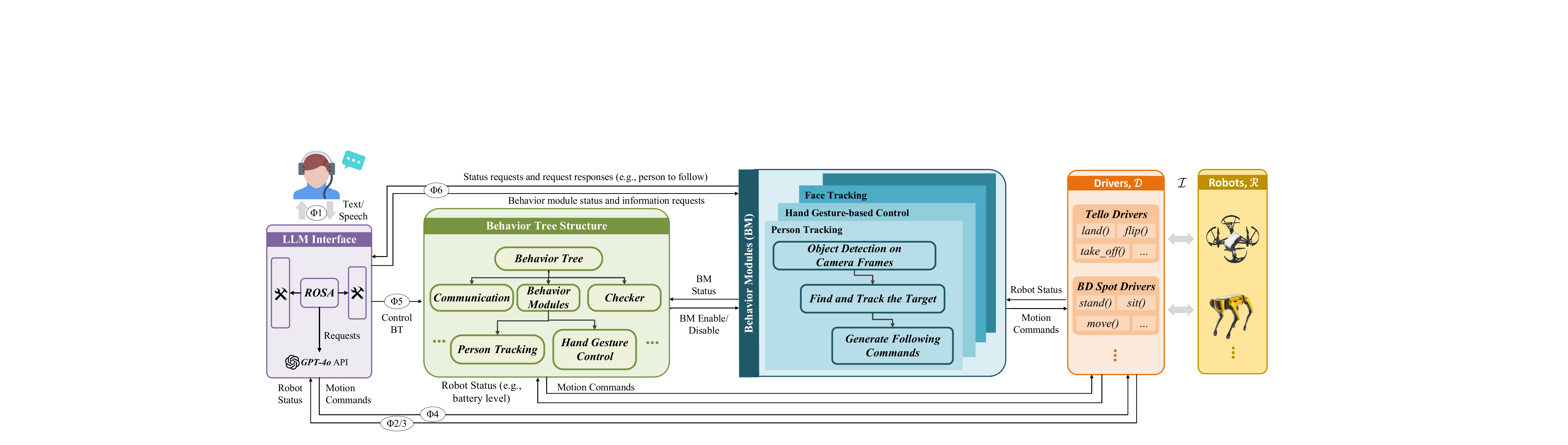}
    \caption{A detailed overview of the proposed system architecture, depicting the integration of LLM-based language understanding, behavior tree core, plugin, and driver modules. Arrow labels indicate the interaction category ($\Phi1$ to $\Phi6$) corresponding to evaluation scenarios described in \S\ref{sec_eval_setup}.}
    \label{fig_architecture}
\end{figure*}
\section{Proposed Framework}
\label{sec_proposed}

The proposed system is a modular and open-source framework publicly accessible via \url{https://github.com/snt-arg/robot\_suite}, featuring an LLM interface built on top of ROSA's \cite{rosa2023} architecture.
It is designed to be robot-agnostic and adaptable to a wide range of robotic tasks and hardware configurations.
As shown in Fig.~\ref{fig_architecture}, the system operates as an end-to-end application with components that can evolve independently, enabling plug-and-play integration of new plugins (\textit{i.e.,} behaviors) without complexity or the need for overall system modifications.

According to the figure, when the user issues natural language commands, they serve as the primary inputs for task interpretation and planning.
These commands reflect the user's intention to initiate specific robot actions, which in turn demand triggering the appropriate control \textit{drivers} (\S\ref{proposed_driver}) tailored for the robot.
The unstructured inputs are then processed by the \textit{LLM interface} module (\S\ref{proposed_llm}), which employs \texttt{GPT-4} backbone to translate them into structured action commands.
At the core of the system, a \textit{behavior tree}(\S\ref{proposed_bt}) handles high-level decision-making procedures based on the input interpreted by the LLM, also offering task execution alongside transparent reasoning.
To support various capabilities, a set of \textit{behavior modules} (\S\ref{proposed_plugin}) (implemented as callable plugins) is integrated into the system.
These behavior modules (\textit{e.g.,} person-following) are independently developed and can be triggered by the BT, operating in conjunction with hardware-specific drivers.
In this regard, the \textit{LLM interface} interacts with both the \textit{behavior tree} (by activating relevant nodes of the tree) and the \textit{behavior modules}, either by querying their status or through bidirectional communication (as in the person-following task, where it requires specifying the target and receiving on-demand updates).
It should be noted that although we present a sample set of executable plugins and the behavior tree structure, the system supports a diverse range of behaviors, making it extensible to employ additional robot capabilities.

\subsection{Robots and Drivers}
\label{proposed_driver}

Interfacing between the high-level decision logic and the physical robots, the \textit{Drivers} module plays a crucial role in executing commands generated by the system (either the LLM interface or the plugins).
These drivers implement atomic, hardware-specific operations (such as \texttt{take\_off}, \texttt{land}, or \texttt{stand}), tailored to the capabilities of different robot platforms.
The driver layer abstracts and encapsulates these low-level controls, enabling straightforward integration across robots.
The current version of the framework supports both \textbf{legged robots}, such as the Boston Dynamics Spot\textsuperscript{\textregistered}, and \textbf{aerial platforms}, including drones like the DJI Tello.
However, it is designed to be easily extensible to additional robot platforms with minimal effort required for integration.
As shown in Fig.~\ref{fig_architecture}, robot-specific drivers provide the interface layer between the physical robot hardware and the rest of the framework, handling bidirectional communications by executing low-level control commands (\textit{e.g.,} motion and velocity) and transmitting robot status feedback (\textit{e.g.,} battery or connectivity).

Consequently, for the supported robots \(\mathbf{R} = \{r_i \mid i \in \mathbb{N}\}\), there exists a mapping \(\mathcal{D} : \mathbf{R} \rightarrow \mathcal{D}_R\) such that each robot \(r_i \in \mathbf{R}\) is associated with a specific driver \(d_{r_i} = \mathcal{D}(r_i)\).
Accordingly, task execution on robot \(r_i\) is delegated to the corresponding driver \(d_{r_i}\) via an interface function \(\mathcal{I} : \mathcal{C} \times \mathbf{R}\), where \(\mathcal{C} = \{c_j \mid j \in \mathbb{N}\}\) is a set of low-level robot commands (\textit{e.g.,} \texttt{land}).
Thus, the appropriate low-level action on \(r_i\) is triggered using:
\begin{equation}
    \mathcal{I}(r_i, c_j) = d_{r_i}(c_j)
    \label{eq_interface}
\end{equation}

\subsection{LLM Interface}
\label{proposed_llm}

\begin{algorithm}[t]
    \caption{LLM Interface Module.}
    \label{alg_llm}
    \scriptsize
    \begin{algorithmic}[1]
        \Procedure{\textbf{Init}}{}
            \State Initialize \texttt{ROS2} node, publishers, and subscribers.
            \State Initialize and configure \texttt{ROSA}.
        \EndProcedure
        \Statex
        \Procedure{\textbf{CommandExecuter}}{}
            \State \textit{query} $\gets$ user input.
            \State Submit \textit{query} to LLM for interpretation and tool selection.
            \If{LLM maps \textit{query} to a valid tool}
                \State Run corresponding tool \Comment{like \texttt{takeoff()}, \texttt{land()}, \texttt{move()}, \textit{etc.}}.
                \State Publish messages to \texttt{ROS2} topics \Comment{robot command, BT update, or plugin response}
            \Else
                \State Inform the user that the command could not be understood/executed.
            \EndIf
        \EndProcedure
        \Statex
        \Procedure{\textbf{Main}}{}
            \State Run \textbf{Init()}
            \If{\textit{cmd} $\gets$ natural language input.}
                \State Run \textbf{CommandExecuter()}
            \EndIf
        \EndProcedure
    \end{algorithmic}
\end{algorithm}

The framework builds upon and extends ROSA \cite{rosa2023} to better support our proposed \ac{BT}-based control structure.
In this regard, Algorithm~\ref{alg_llm} outlines the core logic of the LLM Interface module within the framework.
Upon receiving a natural language input from the user, the module submits it as a prompt to the LLM, interprets the generated result, and dispatches the appropriate behavior by invoking the corresponding robot control tool.
It should be noted that the LLM interface operates as \textbf{one-shot prompting}, where the initial prompt includes contextual instructions, accessible tool descriptions, and behavioral constraints, enabling the LLM to generate outputs without iterative dialogue.

The framework introduces four key extensions to the original ROSA's architecture: \textbf{(i)} autonomous behavior selection, \textbf{(ii)} multimodal HRI support, \textbf{(iii)} failure reasoning, and \textbf{(iv)} structured control integration through \ac{BT}.
A detailed description of each of these new features is presented below: \\

\noindent \textbf{(i) Autonomous behavior selection: }
Unlike frameworks like ROSGPT \cite{rosgpt} and ROSA \cite{rosa2023}, which rely solely on the LLM back-end to interpret given user commands and directly invoke the required tools, our framework introduces an explicit \textit{decision-making layer} for behavior selection.
This layer provides an internal mechanism to autonomously determine which behavior modules (implemented as framework plugins) to trigger based on both the interpreted user intent and additional contextual signals.
Suppose \(\mathcal{B} = \{b_1, b_2, ..., b_n\}\) contains the set of available behavior modules and \(\mathcal{Q} = \{q_1, q_2, ..., q_m\}\) represents the set of semantically parsed queries derived from the LLM interface.
The autonomous selection mechanism chooses the most appropriate behavior candidate for \(q_i \in \mathcal{Q}\) using:
\begin{equation}
    \Gamma : \mathcal{Q} \rightarrow \mathcal{B}
\end{equation}
\noindent which maps each interpreted query $q_i$ to a suitable behavior \(b_i \in \mathcal{B}\) through \(b_{q_i} = \Gamma(q_i)\).
This query-to-behavior mapping is not purely based on simple keyword matching or static prompts.
Instead, $\Gamma(...)$ considers the task context, robot capabilities, and relevant sensor feedback.
For instance, upon receiving the command ``move forward for five seconds," the behavior selection function evaluates runtime conditions, such as battery level or robot operational status, to ensure feasibility before dispatching the appropriate behavior module.
This layered abstraction provides decoupling of \textit{semantic understanding} from \textit{low-level behavior invocation}, offering flexibility and robustness in decision-making.
Finally, once the behavior $b_{q_i}$ is selected, it delegates low-level command execution to the corresponding robot driver module $d_{r_i}$ to invoke the specific robot action $c_j$ through the described interface function $\mathcal{I}(r_i, c_j)$ in Equation~\ref{eq_interface}.\\

\noindent \textbf{(ii) Multimodal HRI support:}
While ROSA supports only text-based interaction with the user, the proposed framework extends the HRI channel by integrating a multimodal input interface \(\mathcal{U} = \{\mathrm{text}, \mathrm{voice}\}\), with each \(u_i \in \mathcal{U}\) serves as a source of natural language commands to the LLM interface.
It should be noted that the current work limits the interface modality to $\mathrm{text}$ (using the keyboard input) to ensure controlled execution and reproducibility.
Nonetheless, the framework's architecture is designed to be readily extensible to other modalities, which requires only adding an appropriate perception module to $\mathcal{B}$ and extending the mapping $\Gamma$ to include the corresponding behaviors.
The multimodal HRI interface functions as an additional pre-processing layer to the LLM interface (\S\ref{proposed_llm}), converting inputs such as $\mathrm{voice}$ into standard $\mathrm{text}$ for the LLM, which is then forwarded for semantic parsing and behavior selection.
Accordingly, it ensures that all input modalities follow a unified \textit{interpretation–decision–execution} pipeline, preserving consistency across heterogeneous HRI channels.
\\


\noindent \textbf{(iii) Failure reasoning:}
The proposed framework extends its baseline capabilities by integrating a failure reasoning and explanation mechanism inspired by \cite{tagliamonte2024generalizable}.
With this, the system has a more advanced task failure management strategy, which invokes an explanation function when a given query \(q_i\) fails to be executed.
When an interpreted query \(q_i \in \mathcal{Q}\) maps to a behavior module \(b_{q_i}\) but fails to achieve a successful execution, the failure analysis stage is triggered to invoke the explanation function, as defined below:
\begin{equation}
    \Psi : \mathcal{F} \rightarrow \mathcal{E}
\end{equation}
\noindent where \(\mathcal{F}\) is a set of identifiable failure modes and \(\mathcal{E}\) represents the set of natural language explanations generated for the user.
The failure reasoning mechanism uses the LLM back-end to transform low-level failure reports of the driver \(d_{r_j}\) or the behavior module \(b_{q_i}\) into semantically-aligned and context-aware reasoning.
Hence, such explanations and structured feedback contribute to the framework's transparency and simplify the HRI procedure. \\

\noindent \textbf{(iv) Structured control integration with \ac{BT}:}
A significant limitation of ROSA is the absence of an explicit structured control logic, which can result in inconsistent execution when handling multi-step or conditional commands.
The proposed framework addresses this by embedding a BT-based architecture at the core of the pipeline, enabling hierarchical and reactive control flow.
Accordingly, each user task \(t_i \in \mathcal{T}\) derived from an interpreted query $q_i$ and its mapped behavior $b_{q_i}$ is encapsulated as a behavior tree node \(n_i\) as below:
\begin{equation}
    n_i \in \mathcal{N} \rightarrow \mathrm{status} \in \{\mathrm{success}, \mathrm{failure}, \mathrm{running}\}
    \label{eq_node_map}
\end{equation}
\noindent where \(\mathcal{N}\) is the set of all active tree nodes.
Accordingly, the behavior tree imposes an execution policy to govern the order and conditional flow between nodes, while ensuring the coherent coordination of behaviors.
This integration allows the BT to dynamically invoke nodes based on LLM interpretations, runtime conditions from the driver interface \(\mathcal{I}(r_i, c_j)\), and sensor feedback, maintaining a continuous alignment between high-level user intent and low-level robotic execution.
The detailed behavior tree composition and node taxonomy are presented in Section~\ref{proposed_bt}.

\subsection{Behavior Tree Structure}
\label{proposed_bt}

The behavior tree represents an executable policy constrained by both the robot's capabilities and the environmental context.
The strength of the behavior tree within the proposed framework lies in its \textit{configurable hierarchical design} and \textit{tick-based execution mechanism}, which enable fine-grained control, flexible task composition, and real-time responsiveness to dynamic system states.
In this regard, the BT operates through discrete execution cycles (or ``\textit{ticks}''), during which the tree is traversed from the root node, evaluating and executing child nodes sequentially from left to right.
Each node returns a \(\mathrm{status}\) (as shown in Algorithm~\ref{eq_node_map}), which determines the subsequent flow of execution within the tree.
This tick-based mechanism enables the behavior tree to continuously update node statuses and adapt the overall behavior in real-time to evolving tasks and conditions.

Let \(\mathcal{S}\) denote the overall task space and \(\Pi\) represent the employed behavior tree in the proposed framework.
Each node \(n_i \in \mathcal{N}\) in the behavior tree is associated with a specific behavior \(b_i \in \mathcal{B}\) implemented as a behavior module, encapsulating its execution logic and up-to-date status.
In this regard, the behavior mapping function is defined as:
\begin{equation}
    \Lambda: \mathcal{S} \times \mathcal{N} \rightarrow \mathcal{B}
\end{equation}
\noindent where a given task \(s_j \in \mathcal{S}\), in combination with the active node \(n_i\), determines the selected behavior \(b_i\).
Thanks to the autonomous behavior selection function $\Gamma(...)$ in the framework, the mapped behavior $b_{q_k}$ inside \(\Pi\) for an interpreted query $q_k$ is obtained as:
\begin{equation}
    b_{q_k} = \Gamma(q_k) = \Lambda(s_j, n_i)
\end{equation}

\subsection{Behavior Modules}
\label{proposed_plugin}

When integrated with the behavior tree \(\Pi\), each behavior module \(b_{q_i}\) of the framework operates as a separate \texttt{ROS2} node that remains inactive until being explicitly triggered by the LLM-interpreted query $q_i$.
Complex behaviors are implemented by coordinating multiple low-level control actions across different subsystems and by feeding their outcomes back into the BT as updated node statuses.
Thanks to the $\Gamma(.)$ and $\Lambda$ functions, the system ensures that behaviors are mapped and executed only when the current task logic determines them to be relevant.
The interaction and tick-based triggering between \(\Pi\) and the behavior modules happens through \texttt{ROS2} services.
Once enabled, the tree ``ticks'' its corresponding behavior module, which typically loads some inputs and produces outputs as part of its execution cycle.

Although behavior modules can continuously receive data through their dedicated communication channels, they can only broadcast system-driven data (such as signals and commands) when triggered by the BT.

Triggering a behavior function can be viewed as an event-driven execution loop with a frequency governed externally by the behavior tree.
This design enforces strict control flow, ensuring that execution aligns with the current task logic.
In the \textit{interpretation–decision–execution} pipeline of the framework, this operation corresponds to the three stages \(\rho \in \{\mathrm{cog}, \mathrm{disp}, \mathrm{exec}\}\), representing LLM-based cognition of the user command, its dispatching to the appropriate behavior module, and the subsequent execution of the selected behavior, respectively.
\section{Experimental Results}
\label{sec_evaluation}

A series of real-world experiments in diverse scenarios was conducted to evaluate the proposed system's capabilities and potential.
In this regard, we designed realistic scenarios, collected our in-house dataset, and evaluated the system’s performance in various tasks.
It is worth noting that the proposed system incorporates a diverse range of behaviors; however, for the experiments, we employed a limited but representative behavior tree structure along with a curated set of sample behavior modules.

\subsection{Evaluation Criteria}
\label{sec_eval_setup}

\begin{table*}[t]
    \centering
    \scriptsize
    \caption{Evaluation scenarios designed to assess the system’s performance across a range of HRI tasks. The expected LLM responses in the rightmost column should remain semantically aligned with the given examples, despite possible variations in wording.} A battery level below $20$\% (\faBatteryQuarter) is considered low, while levels above are treated as sufficient (\faBatteryFull).
    \label{tbl_eval_scenario}
    \begin{tabular}{l|c|c|p{3cm}|p{3cm}|p{5.5cm}}
        \toprule
            \textbf{Category} & \textbf{\#} & \textbf{Platform} & \textbf{User Instruction} & \textbf{Environment Status} &\textbf{Expected System Behavior \& Response} \\
        \midrule
            \textit{Unsupported } ($\Phi1$) & $\Phi1.1$ & \drone/\spot & ``Jump" & \texttt{landed}/\texttt{stand}/\texttt{flying} & Not triggered, response: ``I cannot perform this action." \\
        \midrule
            \multirow{5}{*}{\textit{\shortstack{Context-Aware \\ Response ($\Phi2$)}}} & $\Phi2.1$ & \drone & ``Do a Flip" & \texttt{flying}, \faBatteryFull & Flip maneuver executed. \\
        \cmidrule{2-6}
            & $\Phi2.2$ & \drone & ``Do a Flip" & \texttt{landed}, \faBatteryFull & Blocked due to status, response: ``I cannot do it as the drone is on the ground." \\
        \cmidrule{2-6}
            & $\Phi2.3$ & \drone & ``Do a Flip" & \texttt{landed}, \faBatteryQuarter & Blocked due to battery level and status, response: ``I cannot do it due to low battery and robot status." \\
        \midrule
            \multirow{10}{*}{\textit{\shortstack{System Inquiry\\ Response ($\Phi3$)}}} & $\Phi3.1$ & \drone & ``What is the battery level?" & \texttt{landed}/\texttt{flying}, \faBatteryFull & Response: ``The battery level is \textit{X}\%". \\
        \cmidrule{2-6}
            & $\Phi3.2$ & \drone & ``Can I do a flip with the drone?" & \texttt{flying}, \faBatteryFull & Response: ``Yes, since the drone is flying and the battery level is 26\%". \\
        \cmidrule{2-6}
            & $\Phi3.3$ & \drone & ``Which actions can I perform with the drone?" & \texttt{landed}/\texttt{flying}, \faBatteryFull & Response: ``You can [a list of supported actions]". \\
        \cmidrule{2-6}
            & $\Phi3.4$ & \drone & ``What is the status of the drone?" & \texttt{landed}/\texttt{flying}, \faBatteryFull & Response: ``The drone is on the ground with a battery of 26\%". \\
        \cmidrule{2-6}
            & $\Phi3.5$ & \drone & ``What are the common causes that I get unknown status?" & \texttt{landed}, \texttt{disconnected} & Response: ``The common causes for the robot state being unknown could be [a list of common causes]". \\
        \midrule
            \multirow{4}{*}{\textit{\shortstack{Motion Command \\ ($\Phi4$)}}} & $\Phi4.1$ & \drone/\spot & ``Turn left/right for \texttt{X} sec." & \texttt{stand}/\texttt{flying}, \faBatteryFull & Motion action triggered with duration control. \\
        \cmidrule{2-6}
            & $\Phi4.2$ & \drone/\spot & ``Move forward/backward for \texttt{X} sec (with velocity \texttt{Y})." & \texttt{stand}/\texttt{flying}, \faBatteryFull & Motion action triggered with duration control. \\
        \midrule
        \multirow{5}{*}{\textit{\shortstack{Plugin Switching \\ ($\Phi5$)}}} & $\Phi5.1$ & \drone/\spot & ``Change the control to hand gesture." & \texttt{stand}/\texttt{flying}, \faBatteryFull & Control mode switched, response: ``You can now control the robot using hand gestures." \\
        \cmidrule{2-6}
            & $\Phi5.2$ & \drone/\spot & ``Change the control to keyboard." & \texttt{stand}/\texttt{flying}, \faBatteryFull & Control mode switched, response: ``You can now control the robot using the keyboard." \\
        \midrule
            \multirow{5}{*}{\textit{\shortstack{Vision-based \\ Interaction ($\Phi6$)}}} & $\Phi6.1$ & \drone & ``Track the person with a phone" & \texttt{flying}, \faBatteryFull & Person identified and approached, response: ``Now tracking the person with a phone." \\
        \cmidrule{2-6}
            & $\Phi6.2$ & \drone & ``Track the person with a phone" & \texttt{flying}, \faBatteryFull & Blocked due to identification failure; the drone waits for \(5~\mathrm{sec.}\) to find the person and then halts, response: ``No person with a phone detected" \\
        \bottomrule
    \end{tabular}
\end{table*}

\noindent \textbf{Robot Platform.}
The introduced system is designed to be platform-agnostic, independent of a specific robot type.
This paper demonstrates the system’s functionality on distinct robotic platforms: a lightweight aerial drone (DJI Tello) utilizing its onboard camera, and a legged robot (Boston Dynamics Spot\textsuperscript{\textregistered}) equipped with an external Intel RealSense D435 camera for visual perception.

\noindent \textbf{Software Stack.}
The system is implemented in Python using \texttt{ROS2 Humble} for inter-module communication.
Based on ROSA, it employs the OpenAI API with the \texttt{gpt-4o} model as the \ac{LLM} backend, enabling advanced natural language understanding and reasoning capabilities for interpreting user commands and interacting with behavior modules.
It is worth noting that during the experiments, we relied exclusively on textual input for querying the LLM interface to ensure consistency and facilitate reproducible evaluation.

\noindent \textbf{Dataset and Scenarios.}
We designed diverse scenarios to evaluate the system as an \emph{end-to-end} application, where natural language commands from the user are interpreted and translated into actionable behaviors executed by the robot.
To assess the system's performance in these scenarios, we collected a multi-modal dataset comprising:
\begin{itemize}
    \item Natural language instructions paired with ground-truth behavior annotations (expected responses),
    \item RGB image streams from the robot’s onboard cameras for vision-based perception tasks; and
    \item Execution logs including ``timestamps" and ``active nodes" of the behavior tree for control flow analysis.
\end{itemize}
The mentioned scenarios are summarized in Table~\ref{tbl_eval_scenario}, providing an overview of the task types for evaluation, with their inter-module interactions tagged in Fig.~\ref{fig_architecture}.
Accordingly, all scenarios are designed to evaluate the performance and robustness of the proposed system across diverse real-world \ac{HRI} tasks, covering:
\begin{itemize}
    \item \textit{$\Phi1$ -- Unsupported Actions:} testing the system's ability to reject infeasible commands;
    \item \textit{$\Phi2$ -- Context-Aware Response:} considering the impact of environmental or system state in decision-making;
    \item \textit{$\Phi3$ -- System Inquiry Response:} assessing the system's capacity to answer questions regarding its status, active tasks, or capabilities;
    \item \textit{$\Phi4$ -- Motion Commands:} focusing on low-level action chaining with conditional execution;
    \item \textit{$\Phi5$ -- Plugin Switching:} evaluating dynamic control modality changes of the system; and
    \item \textit{$\Phi6$ -- Vision-based Interactions:} investigating the system's perception-driven behavior modules based on visual attributes.
\end{itemize}

\noindent \textbf{Evaluation Metrics.}
Since each evaluation scenario embodies distinct objectives and interaction modalities, we adopt both ``qualitative" and ``quantitative" metrics tailored to the specific characteristics of each task.
These metrics are designed to assess the system's performance in terms of response clarity and correctness, perceptual accuracy, and successful execution of intended behaviors.
The evaluation criteria employed in this study are summarized as follows:

\begin{itemize}
    \item \textit{Success Rate:} the degree to which the system achieves the intended task outcome across scenarios \(\Phi1\) to \(\Phi6 \).
    Given the variability in LLM responses, we adopt a weighted success strategy, where each scenario case is evaluated over \(k=10\) independent runs and across the introduced stages \(\rho\).
    Each demanded behavior is scored using a binary metric \( \eta^\rho_{i,j} \in \{0, 1\} \), where $i$ indexes the scenario case and $j$ the iteration, with score $1$ denoting success and $0$ denoting failure.
    Thus, success rate is calculated as: \[ \sigma_i = \frac{1}{k} \sum_{j=1}^{k} \left( \eta_{i,j}^{\text{cog}} + \eta_{i,j}^{\text{disp}} + \eta_{i,j}^{\text{exec}} \right) \]
    \item \textit{Response Latency:} the time delay between the start and end timestamps of stage $\rho$, i.e., \(t^\rho_s\) and \(t^\rho_e\). For each scenario case, latency is measured as \(\mathcal{L}_\rho = t^\rho_e - t^\rho_s \), and the total latency will be \(\mathcal{L}_{total} = \sum_{i=1}^{\rho} \mathcal{L}_i \).
\end{itemize}

It should be noted that explicitly benchmarking the performance of underlying LLM and computer vision models is beyond the current scope of this paper.
The primary objective of the proposed framework is to remain \textbf{model-agnostic} and to support any standard LLM or vision model that provides an accessible API for request–response communication.
In this design, the LLM and perception components act as interchangeable \textit{back-end modules} responsible for language understanding and environmental perception, respectively.
This modular architecture enables the integration of other models without requiring modifications to the overall system structure.
While the selection of a particular LLM (such as \texttt{OpenAI GPT} in our implementation) may influence factors such as reasoning quality or response latency, our focus in this work is on demonstrating the end-to-end functionality, scalability, and generalizability of the framework. A systematic comparison of alternative LLM back-ends and their influence on overall system performance will be explored in future work.
\subsection{Benchmarking Setup}
\label{sec_eval_implementation}

\begin{figure}[t]
    \centering
    \includegraphics[width=.8\columnwidth]{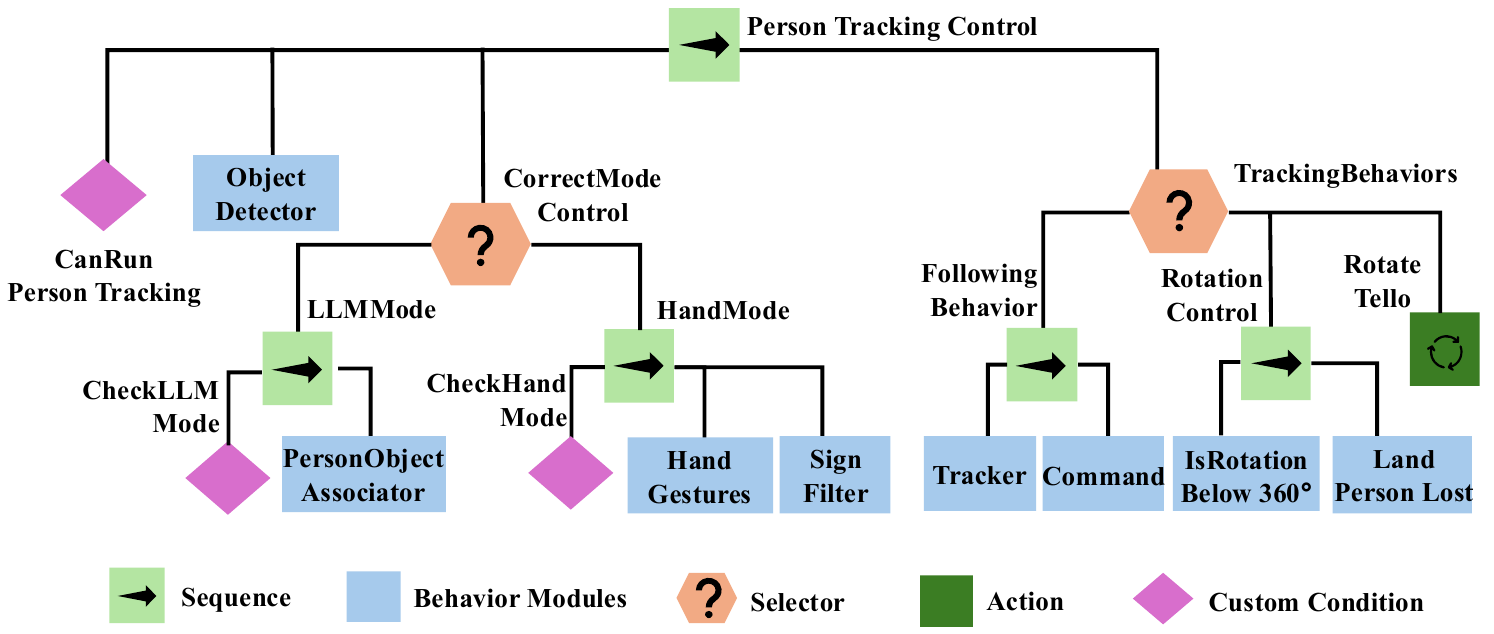} 
    \caption{Structure of the sample behavior tree employed in the paper for system evaluation, containing the hierarchical arrangement of execution nodes to manage robotic behaviors.}
    \label{fig_bt}
\end{figure}

\noindent \textbf{Custom behavior tree structure.}
The behavior tree shown in Fig.~\ref{fig_bt} serves as a representative implementation used for evaluation.
However, the framework is agnostic to the specific BT structure and can be extended to any compatible design.
The behavior tree employed in our benchmarking setup comprises multiple node types, which can be categorized into \textit{control nodes} and \textit{custom nodes}.
While the control nodes are responsible for governing the execution flow, the custom nodes handle task-specific robot behaviors that are aligned with the framework's architecture.

\noindent \textbf{Custom behavior modules.}
Similar to the behavior tree structure, the framework can support a diverse set of behavior modules, each encapsulating a specific perceptual or control capability.
As depicted in Fig.~\ref{fig_bt}, the current paper benchmarks the framework's capabilities using two representative behavior modules as a proof of concept: \textit{person tracking} and \textit{hand gesture-based control}.

The \textit{hand gesture-based control} enables intuitive robot control through hand gestures, allowing users to issue behavioral commands using predefined gesture patterns.
Once triggered, it processes real-time gesture inputs to dynamically control the robot’s actions.
This modality offers a responsive interface, particularly useful in scenarios where text-driven control is unavailable.
The \textit{person tracking} behavior module enables the robot to identify and track a person in real-time by processing the video feed from the robot’s onboard/mounted camera and applying an object detection algorithm to identify individuals in the scene.
To implement this behavior module, we have employed YOLO11 \cite{yolov11} for real-time object detection.
As summarized in Algorithm~\ref{alg_tracking}, detected objects are passed to the LLM, which determines the appropriate target for tracking based on the user intent.
Once a target person is selected, the plugin continuously issues movement commands to the robot via the driver interface to maintain focus on and follow the individual.

\begin{algorithm}[t]
    \caption{Person Tracking Behavior Module.}
    \label{alg_tracking}
    \scriptsize    
    \begin{algorithmic}[3]
        \Procedure{\textbf{Init}}{}
            \State Initialize drivers, plugins, LLM, and BT
            \State \textit{person} $\gets$ \textit{null}
        \EndProcedure
        \Statex
        \Procedure{\textbf{Track}}{}
            \State Send \textit{bounding\_boxes} to LLM Interface
            \State \textit{tracking\_signal} $\gets$ LLM Interface
            \If{\textit{tracking\_signal} equals \texttt{"tracking"}}
                \State \textit{person} $\gets$ \textit{bounding\_boxes} overlapping \textit{tracking\_signal}
                \State \textit{vel\_cmd} $\gets$ Build velocity command from \textit{person}
                \State Publish \textit{vel\_cmd} topic to robot
            \EndIf
        \EndProcedure
        \Statex
        \Procedure{\textbf{Lost}}{}
            \State \textit{direction} $\gets$ last known \textit{person} position
            \State \textit{vel\_cmd} $\gets$ rotate in axis in \textit{direction}
            \State Publish \textit{vel\_cmd} topic to robot \Comment{Rotate to identify the \textit{person}}
        \EndProcedure
        \Statex
        \Procedure{\textbf{Main}}{}
            \State Run \textbf{Init()}
            \State \textit{tracking\_signal} $\gets$ \textit{null}
            \ForAll{\textit{frame} of \textit{camera frames}}
                \State \textit{bounding\_boxes} $\gets$ detected \texttt{person} objects in \textit{frame} using \textbf{YOLO}
                \If{\textit{bounding\_boxes} is not \textit{null}}
                    \State Run \textbf{Track()}
                \Else
                    \State Run \textbf{Lost()}
                \EndIf
            \EndFor
        \EndProcedure
    \end{algorithmic}
\end{algorithm}



\subsection{Success Rate Analysis}
\label{sec_eval_sr}

\begin{table}[t]
    \centering
    \caption{Success Rate ($\sigma$) of behavior execution across scenarios over \(k=10\) independent runs. $\eta_i^{\text{cog}}$, $\eta_i^{\text{disp}}$, and $\eta_i^{\text{exec}}$ refer to LLM cognition, dispatching stage, and task execution success levels of the scenario $s_i$, respectively.}
    \label{tbl_eval_sr}
    \begin{tabular}{c|c|c|c|c|c|l}
        \toprule
            \textbf{\#} & \textbf{Robot} & \textbf{$\eta^{\text{cog}}$} & \textbf{$\eta^{\text{disp}}$} & \textbf{$\eta^{\text{exec}}$} & \textbf{$\sigma$} & \textbf{Details} \\
        \midrule
            $\Phi$1.1 & \drone & \cellcolor{greend}{\(1.0\)} & \(-\) & \cellcolor{greend}{\(1.0\)} & \cellcolor{greend}{\(1.00\)} & \\
            $\Phi$1.1 & \spot & \cellcolor{greend}{\(1.0\)} & \(-\) & \cellcolor{greend}{\(1.0\)} & \cellcolor{greend}{\(1.00\)} & \\
        \midrule
            $\Phi$2.1 & \drone & \cellcolor{greend}{\(1.0\)} & \cellcolor{yellow}{\(0.7\)} & \cellcolor{greenl}{\(0.8\)} & \cellcolor{greenl}{\(0.83\)} & {Flip direction varies with input} \\
            $\Phi$2.2 & \drone & \cellcolor{greend}{\(1.0\)} & \cellcolor{greend}{\(1.0\)} & \cellcolor{greenl}{\(0.8\)} & \cellcolor{greend}{\(0.93\)} & {Flip direction varies with input} \\
            $\Phi$2.3 & \drone & \cellcolor{greend}{\(1.0\)} & \cellcolor{greend}{\(1.0\)} & \cellcolor{greend}{\(1.0\)} & \cellcolor{greend}{\(1.00\)} & \\
        \midrule
            $\Phi$3.1 & \drone & \cellcolor{greend}{\(1.0\)} & \cellcolor{greend}{\(1.0\)} & \cellcolor{greend}{\(1.0\)} & \cellcolor{greend}{\(1.00\)} \\
            $\Phi$3.2 & \drone & \cellcolor{greend}{\(0.9\)} & \cellcolor{greenl}{\(0.8\)} & \cellcolor{greend}{\(1.0\)} & \cellcolor{greend}{\(0.90\)} \\
            $\Phi$3.3 & \drone & \cellcolor{greend}{\(1.0\)} & \(-\) & \cellcolor{greend}{\(1.0\)} & \cellcolor{greend}{\(1.00\)} \\
            $\Phi$3.4 & \drone & \cellcolor{greend}{\(1.0\)} & \cellcolor{greend}{\(1.0\)} & \cellcolor{greend}{\(1.0\)} & \cellcolor{greend}{\(1.00\)} \\
            $\Phi$3.5 & \drone & \cellcolor{greend}{\(1.0\)} & \(-\) & \cellcolor{greend}{\(1.0\)} & \cellcolor{greend}{\(1.00\)} \\
        \midrule
            $\Phi$4.1 & \drone & \cellcolor{yellow}{\(0.7\)} & \cellcolor{greend}{\(1.0\)} & \cellcolor{yellow}{\(0.7\)} & \cellcolor{greenl}{\(0.80\)} & {Velocity defaults without input} \\
            $\Phi$4.1 & \spot & \cellcolor{yellow}{\(0.7\)} & \cellcolor{greend}{\(1.0\)} & \cellcolor{yellow}{\(0.7\)} & \cellcolor{greenl}{\(0.80\)} & {Velocity defaults without input} \\
            $\Phi$4.2 & \drone & \cellcolor{greenl}{\(0.8\)} & \cellcolor{greend}{\(1.0\)} & \cellcolor{greenl}{\(0.8\)} & \cellcolor{greenl}{\(0.87\)} & {Velocity defaults without input} \\
            $\Phi$4.2 & \spot & \cellcolor{yellow}{\(0.7\)} & \cellcolor{greend}{\(1.0\)} & \cellcolor{yellow}{\(0.7\)} & \cellcolor{greenl}{\(0.80\)} & {Velocity defaults without input} \\
        \midrule
            $\Phi$5.1 & \drone & \cellcolor{greend}{\(1.0\)} & \cellcolor{greend}{\(1.0\)} & \cellcolor{greend}{\(1.0\)} & \cellcolor{greend}{\(1.00\)} & {Highly similar LLM outputs} \\
            $\Phi$5.1 & \spot & \cellcolor{greenl}{\(0.8\)} & \cellcolor{greend}{\(1.0\)} & \cellcolor{greenl}{\(0.8\)} & \cellcolor{greenl}{\(0.87\)} & {Highly similar LLM outputs} \\
            $\Phi$5.2 & \drone & \cellcolor{greend}{\(1.0\)} & \cellcolor{greend}{\(1.0\)} & \cellcolor{greend}{\(1.0\)} & \cellcolor{greend}{\(1.00\)} \\
            $\Phi$5.2 & \spot & \cellcolor{greend}{\(0.9\)} & \cellcolor{greend}{\(1.0\)} & \cellcolor{greend}{\(1.0\)} & \cellcolor{greend}{\(0.97\)} & \\
        \midrule
            $\Phi$6.1 & \drone & \cellcolor{greend}{\(1.0\)} & \cellcolor{greend}{\(1.0\)} & \cellcolor{greend}{\(1.0\)} & \cellcolor{greend}{\(1.00\)} \\
            $\Phi$6.2 & \drone & \cellcolor{greend}{\(1.0\)} & \cellcolor{greend}{\(1.0\)} & \cellcolor{greend}{\(1.0\)} & \cellcolor{greend}{\(1.00\)} \\
        \midrule
            \textbf{Avg.} & Any & \cellcolor{greend}{\(\mathbf{0.93}\)} & \cellcolor{greend}{\(\mathbf{0.92}\)} & \cellcolor{greend}{\(\mathbf{0.95}\)} & \cellcolor{greend}{\(\mathbf{0.94}\)} \\
        \bottomrule
    \end{tabular}
\end{table}

The objective of these experiments is to assess the capability of the LLM Interface to generate semantically appropriate interpretations of user commands, dispatch them to proper modules or executors, and verify the correctness of the corresponding task executions.
In this regard, command interpretation accuracy $\eta_i^{\text{cog}}$ for the scenario index $i$ is evaluated through manual analysis to ensure that the system’s understanding of each command aligns with its intended meaning.
Dispatch accuracy $\eta_i^{\text{disp}}$ is assessed by verifying whether the appropriate module or execution pathway is correctly activated in response to the interpreted command.
Finally, $\eta_i^{\text{exec}}$ is determined by observing whether the robot completes the intended task in the given environment, as confirmed by visual observations and system feedback logs.
As illustrated in Table~\ref{tbl_eval_sr}, the results validate the viability of the proposed framework across various real-world scenarios.
The few cases with partially lower \(\sigma\) values are due to cognitive misunderstanding or practical integration issues.

According to the table, the system achieves an average accuracy of $0.93$ for cognition, $0.92$ for task dispatching, $0.95$ for execution, and an overall end-to-end success rate of $0.96$ across different robots and various real-world scenarios.
Out of the $20$ evaluated scenarios, eleven achieved a perfect success rate ($\sigma = 1.00$) and five obtained $\sigma \geq 0.85$, indicating that the LLM was able to correctly interpret the instruction, trigger the appropriate behavior, and observe the expected result.
The interpretation accuracy $\eta_i^{\text{cog}}$ remained near-perfect in most cases, except for the \textit{Motion Command} scenarios $(\Phi4)$ where the motion direction and duration need to be appropriately understood by the LLM.
The legged robot showed lower accuracy than the drone for \textit{Plugin Switching} $(\Phi5)$, possibly due to its complex locomotion dynamics and the higher precision required for behavior activation.
Additionally, in scenario $\mathrm{\Phi3.2}$ under \textit{System Inquiry Response}, the LLM exhibited mild signs of behavior mapping ambiguity, possibly due to context mismanagement.
Dispatching accuracy $\eta_i^{\text{disp}}$ are consistently high across supported actions except from $\mathrm{\Phi2.1}$ and $\mathrm{\Phi3.2}$, where incomplete routing to executors were observed.
Scenarios marked with a dash in the mentioned column indicate cases where no dispatching occurred and the execution took place directly at the driver's level, without requiring the tools to be called.
Finally, regarding $\eta_i^{\text{exec}}$, the rates strongly correlate with the system's state awareness and the level of contextual information required for successful task completion.
This explains why the drone exhibits lower $\eta_i^{\text{exec}}$ in $(\Phi2)$ and $(\Phi4)$, which specifically evaluate motion control and contextual understanding.

In summary, the experimental results demonstrate that the proposed framework maintains a high level of reliability and semantic consistency across heterogeneous robotic platforms and diverse interaction scenarios.
Scenarios with lower $\sigma$ confirm that a clear trend is highly observable between cognition and execution success, where lower $\eta_i^{\text{cog}}$ values mainly correspond to higher reasoning ambiguity or context misinterpretation.
This suggests that the quality of the underlying language model’s semantic understanding is the primary factor influencing the system's end-to-end performance.
\subsection{Runtime Performance}
\label{sec_eval_time}

\begin{table}[t]
    \centering
    \caption{Timing results for selected scenarios, including LLM response time ($\mathcal{L}_{\text{cog}}$), dispatching time ($\mathcal{L}_{\text{disp}}$), and task execution duration ($\mathcal{L}_{\text{exec}}$), in milliseconds ($\mathrm{ms}$).}
    \label{tbl_eval_time}
    \begin{tabular}{c|c|ccc|c|l}
        \toprule
            \textbf{\#} & \textbf{Robot} & \textbf{$\mathcal{L}_{\text{cog}}$} & \textbf{$\mathcal{L}_{\text{disp}}$} & \textbf{$\mathcal{L}_{\text{exec}}$} & \textbf{$t_i^{\text{total}}$} & \textbf{Details} \\
        \midrule
            $\Phi$1.1 & \drone & \cellcolor{yellow}{\(4200\)} & \cellcolor{greend}{\(0\)} & \(-\) & \(4200\) & \\
            $\Phi$1.1 & \spot & \cellcolor{yellow}{\(4000\)} & \cellcolor{greend}{\(0\)} & \(-\) & \(4000\) & \\
        \midrule
            $\Phi$2.1 & \drone & \cellcolor{yellow}{\(4113\)} & \cellcolor{greend}{\(\leq 1\)} & \cellcolor{greend}{\(611\)} & \(4724\) & {LLM may ask for direction} \\
            $\Phi$2.2 & \drone & \cellcolor{yellow}{\(2022\)} & \cellcolor{greend}{\(551\)} & \cellcolor{greenl}{\(1195\)} & \(3768\) & \\
            $\Phi$2.3 & \drone & \cellcolor{yellow}{\(2681\)} & \cellcolor{greend}{\(\leq 1\)} & \(-\) & \(2681\) & \\
        \midrule
            $\Phi$3.1 & \drone & \cellcolor{yellow}{\(2131\)} & \cellcolor{greend}{\(\leq 1\)} & \(-\) & \(2131\) & \\
            $\Phi$3.2 & \drone & \cellcolor{yellow}{\(2540\)} & \cellcolor{greend}{\(\leq 1\)} & \(-\) & \(2540\) & \\
            $\Phi$3.3 & \drone & \cellcolor{yellow}{\(7782\)} & \cellcolor{greend}{\(\leq 1\)} & \(-\) & \(7782\) & {Listing actions takes time} \\
            $\Phi$3.4 & \drone & \cellcolor{yellow}{\(1688\)} & \cellcolor{greend}{\(\leq 1\)} & \(-\) & \(1688\) & \\
            $\Phi$3.5 & \drone & \cellcolor{orange}{\(+9999\)} & \cellcolor{greend}{\(\leq 1\)} & \(-\) & \(+9999\) & {Listing actions takes time} \\
        \midrule
            $\Phi$4.1 & \drone & \cellcolor{yellow}{\(2559\)} & \cellcolor{greenl}{\(1010\)} & \cellcolor{yellow}{\(2321\)} & \(5890\) & {LLM can ask for more info} \\
            $\Phi$4.1 & \spot & \cellcolor{yellow}{\(3675\)} & \cellcolor{greenl}{\(1920\)} & \cellcolor{yellow}{\(3154\)} & \(8749\) & {Re-execution due to LLM} \\
            $\Phi$4.2 & \drone & \cellcolor{yellow}{\(5572\)} & \cellcolor{greend}{\(542\)} & \cellcolor{greenl}{\(1647\)} & \(7761\) & {LLM can ask for more info} \\
            $\Phi$4.2 & \spot & \cellcolor{yellow}{\(2424\)} & \cellcolor{greend}{\(289\)} & \cellcolor{yellow}{\(2713\)} & \(5426\) & {Slower motion execution} \\
        \midrule
            $\Phi$5.1 & \drone & \cellcolor{yellow}{\(2391\)} & \cellcolor{greend}{\(\leq 1\)} & \(-\) & \(2391\) & {Depends on control change} \\
            $\Phi$5.1 & \spot & \cellcolor{yellow}{\(3150\)} & \cellcolor{greend}{\(\leq 1\)} & \(-\) & \(3150\) & {Depends on control change} \\
            $\Phi$5.2 & \drone & \cellcolor{yellow}{\(3023\)} & \cellcolor{greend}{\(\leq 1\)} & \(-\) & \(3023\) & \\
            $\Phi$5.2 & \spot & \cellcolor{yellow}{\(2981\)} & \cellcolor{greend}{\(\leq 1\)} & \(-\) & \(2981\) & \\
        \midrule
            $\Phi$6.1 & \drone & \cellcolor{yellow}{\(2380\)} & \cellcolor{greend}{\(294\)} & \cellcolor{orange}{\(+9999\)} & \(+9999\) & {Varies with user input} \\
            $\Phi$6.2 & \drone & \cellcolor{yellow}{\(2030\)} & \cellcolor{orange}{\(5000\)} & \(-\) & \(7030\) & {Varies with user input} \\
        \bottomrule
    \end{tabular}
\end{table}

The elapsed time for each scenario, comprising LLM interpretation ($\mathcal{L}_{\text{cog}}$), command dispatch ($\mathcal{L}_{\text{disp}}$), and task execution ($\mathcal{L}_{\text{exec}}$), is presented in Table \ref{tbl_eval_time}.
As observed, $\mathcal{L}_{\text{disp}}$ incurs the lowest computational cost among the three stages, with negligible latency in most scenarios ($13$ out of twenty cases with $\mathcal{L}_{\text{disp}} \leq 1$).
While such low values correspond to direct decision-making, scenarios with higher dispatch times typically result from command-specified execution durations embedded in the dispatching process.
The most computationally demanding scenario is $\Phi6.2$, where the user requests person tracking but the camera fails to detect the target.
In this case, the dispatching phase lasts approximately five seconds while the robot attempts to locate the person, after which it halts.
Other time-intensive cases arise in $\Phi4.1$ and $\Phi4.2$ under \textit{Motion Command}, where the system must verify the robot’s status regardless of its type.
Thus, an overhead is likely due to the additional sensing, state validation, and decision-making steps required prior to motion commands.

In most scenarios, the majority of the time is spent querying the LLM in $\mathcal{L}_{\text{cog}}$, which can vary depending on the task complexity and the length of the generated text.
In other words, $\mathcal{L}_{\text{cog}}$ dominates the overall response time due to the time required by the LLM to generate reasonable responses.
Notably, $\Phi3.5$ requires a longer time due to the amount of information that must be displayed to the user.
Regarding $\mathcal{L}_{\text{exec}}$, most actions require the standard execution time of the platform.
The highest execution time can be seen in $\Phi6.1$, where the robot remains in an infinite loop while following the person.
Moreover, \textit{Motion Command} scenarios in $\Phi4.\mathrm{X}$ exhibit great execution time due to multiple factors, including the magnitude of the requested motion value (\textit{e.g.,} distance or altitude) and hardware-specific capabilities.
While the first factor directly affects how long the robot must move before completing the task, the latter is due to the robot’s locomotion variation.
For instance, the legged robot is inherently slower than the drone, resulting in longer completion times.

In general, despite the inherent variability in LLM processing, the system consistently exhibits stable timing behavior and minimal dispatch overhead, demonstrating the efficiency of the $\mathrm{ROS}$-based communication layer and the modular system architecture.
Additionally, scenarios with higher cognition times are primarily associated with task complexity and context-dependent reasoning, where the LLM needs to process longer or more ambiguous instructions.
This is not directly aligned with the $\mathcal{L}_{\text{exec}}$ results, as the execution time is mainly influenced by the robot’s locomotion characteristics and platform-specific speed.
\subsection{Qualitative Results}
\label{sec_eval_qual}

\begin{figure*}[t]
    \centering
    \includegraphics[width=\textwidth]{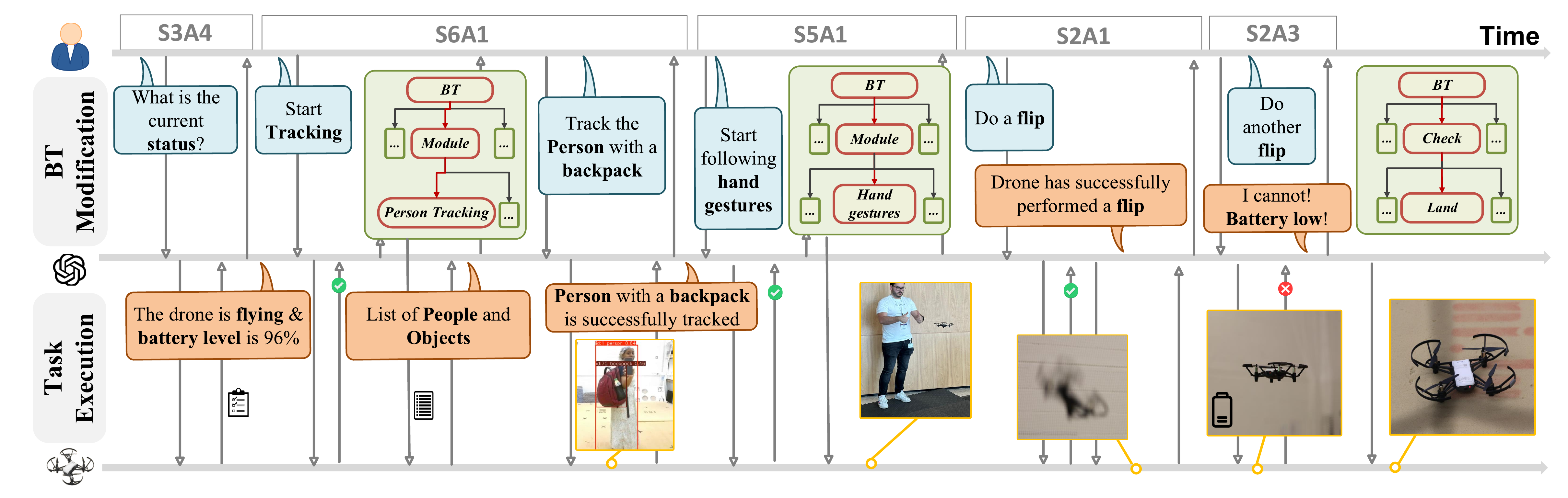}
    \caption{High-level state-flow diagram of the proposed end-to-end \ac{HRI} system, highlighting command interpretation and execution states.}
    \label{fig_eval_object}
\end{figure*}

Fig.~\ref{fig_eval_object} presents a high-level state flow of scenario-based \ac{HRI} evaluations, depicting the transitions and execution states involved in task interpretation and control.
Accordingly, based on the commands given by the user, the task execution is triggered, guiding the system through the appropriate states and transitions.
The figure illustrates the dynamic nature of multimodal HRI, supported by our proposed framework, showing how natural language inputs activate behavior modules and influence the flow of execution.
One dialog is related to the human and the other to the LLM's response to the corresponding task.
\subsection{Discussions}
\label{sec_eval_discussion}

The conducted experiments showcase the capability of the proposed framework as an end-to-end application, integrating an LLM to trigger diverse tasks through its behavior-driven architecture.
By integrating an LLM as the cognitive interface, the framework enables high-level task interpretation and coordination of behavior modules, validating its functionality in real-world HRI settings.
The system performs reliably in scenarios characterized by low to moderate cognitive and physical complexity.
In such cases, the LLM effectively interprets user instructions and triggers the appropriate behavior tree nodes for execution.
However, the execution accuracy relies heavily on the clarity of user commands, and ambiguity or under-specified input (\textit{e.g.}, “Do a flip” without indicating direction) can lead to inconsistent or incomplete behavior triggering.
In these cases, the LLM may either issue a request for clarification or default to predefined assumptions, potentially deviating from user intent.
In more complex scenarios, particularly those requiring multi-step task decomposition, the framework relies on the LLM interface’s ability to reason, formulate a multi-step plan, and activate appropriate behavior modules.

While the system shows promising results in executing demanded tasks, several performance limitations remain.
Performance bottlenecks are primarily linked to the LLM's nature and its dependency on third-party perception or control plugins (such as YOLO11 for person tracking).
These components introduce variability in behavior outcomes, particularly in cases involving sensor noise, uncertain scene dynamics, or subtle linguistic variations in user input.

\section{Conclusions}
\label{sec_conclusions}

This paper presented an open-source, modular, and extensible framework that combines LLMs with behavior trees, enabling robots to interpret and execute natural language commands in a transparent and scalable manner.
The core contributions of this work lie in simplifying human-robot interaction and the interpretability of robot decision-making, enabling more dynamic and adaptable robot behaviors.
Real-world experiments across diverse environments have shown that the system demonstrates a high success rate in both the ``semantic interpretation'' of user instructions and the ``execution of corresponding robotic behaviors'' stages.
These experiments were structured into defined scenarios, encompassing perception-driven tasks and motion commands, to evaluate the system’s robustness and versatility under realistic conditions.

Future work will address system design aspects not covered in this paper.
First, we plan to benchmark the impact of various LLM back-ends (e.g., \texttt{OpenAI}, \texttt{ST}, \texttt{Llama}) and perception-driven behavior modules on the system’s decision-making and execution quality.
Additionally, extending the framework to multi-robot coordination is another step to evaluate collaborative task execution across heterogeneous robots.
Finally, integrating more input modalities and supporting dynamic behavior selection through the BT are additional extensions that can further enhance the framework.


\bibliographystyle{IEEEtran}
\bibliography{root}


\end{document}